# Automated Curriculum Learning for Embodied Agents: A Neuroevolutionary Approach

**Nicola Milano and Stefano Nolfi**
Institute of Cognitive Science and Technologies,
National Research Council, Roma, Italy
nicola.milano@cnr.it, stefano.nolfi@cnr.it

**Abstract**
We demonstrate how an evolutionary algorithm can be extended with a curriculum learning process that selects automatically the environmental conditions in which the evolving agents are evaluated. The environmental conditions are selected so to adjust the level of difficulty to the ability level of the current evolving agents and so to challenge the weaknesses of the evolving agents. The method does not require domain knowledge and does not introduce additional hyperparameters. The results collected on two benchmark problems, that require to solve a task in significantly varying environmental conditions, demonstrate that the method proposed outperforms conventional algorithms and generates solutions that are robust to variations.

## 1. Introduction

Neuroevolution (Lehman and Miikkulaien, 2013) is a method widely used to evolve embodied and situated agents. Clearly, the conditions in which the agents are evaluated affect the course of the evolutionary process. Ideally, the environmental conditions should match the skill level of the evolving agents, i.e. should be sufficiently difficult to exhort an adequate selective pressure and sufficiently simple to ensure that random variations can occasionally produce progresses. This can be obtained by varying the evaluation conditions during the evolutionary process, i.e. by increasing the complexity of the environmental conditions across generations and by selecting conditions that are challenging for the current evolving agents.

A possible method that can be used to achieve this objective is incremental evolution, i.e. the utilization of an evolutionary process divided into successive phases of increasing complexity. For example the evolution of an ability to visually track target objects can be realized by exposing the evolving agents first to large immobile targets, then to small immobile targets, and finally to small moving targets (Harvey, Husband and Cliff, 1994). Similarly, the evolution of agents evolved for the ability to capture escaping prey can be organized in a series of subsequent phases in which the speed of the prey and the pursuit delay is progressively increased (Gomez & Miikkulainen, 1997). However, these approaches presuppose that the tasks can be ordered by difficulty, when in reality they might vary along multiple axes of difficulty. Alternatively, the incremental process can be realized by dividing the problem in a set of sub-problems and by using a multi-objective optimization algorithm that select the agents that excel in at least one sub-problem (Mouret & Doncieux, 2008). In general, incremental approaches can be effective but introduce hard to tune hyperparameters and require the utilization of domain dependent knowledge.

A second possible method consists in competitive co-evolution, i.e. the evolution of agents that compete with other evolving agents. For example, the co-evolution of two populations of predators and prey robots selected for the ability to capture prey and escape predators, respectively (Rosin & Belew, 1997; Nolfi & Floreano, 1998; De Jong, 2005; Chong, Tiňo & Yao, 2009; Miconi, 2009; Samothrakis et. al., 2013; Simione & Nolfi, in press). Indeed, competitive co-evolution can produce automatically a progressive complexification of the environmental conditions for both populations. Moreover, competitive co-evolution permits to expose the evolving agents to conditions that challenge their weaknesses thanks to the fact that the exploitation of their weaknesses has an adaptive value for the opponent population. Unfortunately, competitive co-evolution does not necessarily lead to a progressive complexification of the adaptive problem (for a discussion see Miconi, 2009; Simione and Nolfi, in press). Moreover, it can only be applied to problems that can be formulated in a competitive manner.

A third possible method, that we investigate in this article, consists in enhancing the evolutionary algorithm with a process capable to select the environmental conditions that have the right level of difficulty for the current evolving agents and that challenge the weaknesses of the current evolving agents. We refer to this class of methods with the term curriculum learning.

The utility of curriculum learning for supervised learning has been widely demonstrated and constitutes an active research field (Bengio et al., 2009; Sutskever & Zaremba, 2014; Graves et al., 2016 and 2017; Held et al., 2017). Instead, the usage of curriculum learning in evolutionary methods was not sufficiently explored yet. In a recent work Wang et al. (2019) proposed an algorithm that evolve a population of increasingly complex agent-environmental couples. Each couple is formed by an agent evolved for the ability to operate effectively in certain environmental conditions and by a description of that environmental conditions, i.e. the environmental condition in which the agent is evaluated. The progressive complexification is obtained by: (i) optimizing agents in their associated environments, (ii) generating new environments by creating copies with variation of the existing environments, and (iii) attempting to transfer copies of the current agents into another agent-environmental couple. As shown by the authors, this method permits to produce agents capable to operate effectively in remarkably complex conditions. Moreover, it permits to generate solutions for environmental conditions that results too hard for a conventional evolutionary method. On the other hand, this method produces solutions tailored to specific environmental conditions that do not necessarily generalize in other conditions.

In this work we propose a curriculum learning method that automatically selects the environmental conditions in which the evolving agents are evaluated. The environmental conditions are selected so to adjust the level of difficulty to the ability level of the current evolving agents and so to challenge the weaknesses of the evolving agents. The method can be used in combination with any evolutionary algorithm and does not introduce additional hyperparameters. The results collected on two benchmark problems, that requires to solve a task in significantly varying environmental conditions, demonstrate that the method proposed produces significantly better performance than a conventional method and generates solutions that are robust to variations.

## 2. The method

The rationale of the method consists in estimating the difficulty level of the environmental conditions encountered by the evolving agents of recent generations and in using this information to expose the agents of successive generations to environmental conditions that have the appropriate level of difficulty. The difficulty level of the environmental conditions can be estimated on the basis of the inverse of the relative fitness obtained by the agents evaluated in those conditions. Notice that this also permits to select environmental conditions that threaten the specific weaknesses of the current evolving agents. Indeed, the level of difficulty calculated in this way rate the level of difficulty for the current evolving agents.

We describe our method and the problems used to verify its efficacy in the following sub-sections.

### 2.1 The adaptive problems

To verify the efficacy of the method we selected two problems, commonly used as benchmark in continuous control optimization, that are sufficiently difficult to appreciate the relative efficacy of different algorithms and that require to solve a problem in significantly varying conditions.

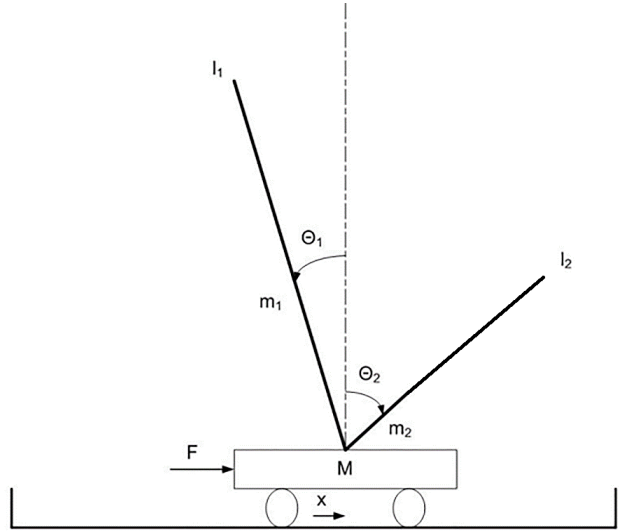

Figure 1. The long double-pole balancing problem.

The first problem is the **Long double-pole balancing** problem (Pagliuca, Milano and Nolfi, 2018; https://github.com/snolfi/longdpole), a harder version than the classic non-markovian double-pole balancing problem (Wieland, 1991)

The problem consists in controlling a cart with two poles, attached with passive hinge joints on the top side of the cart, for the ability to keep the poles balanced (Figure 1, left). The cart has a mass of 1 Kg. The long pole and the short pole have a mass of 0.5 and 0.25 Kg and a length of 1.0 and 0.5 m, respectively. The agent has three sensors, that encode the current position of the cart on the track (x) and the current angle of the two poles ($\theta_1$ and $\theta_2$) and one motor. The activation state of the motor is normalized in the range [-10.0, 10.0] N and is used to set the force applied to the cart.

The neural network controller of the agent is constituted by a LSTM network (Gers and Schmidhuber, 2001) with 3 layers, 3 sensory neurons, 10 internal units, and 1 motor neuron. We used a LSTM network since the problem requires to determine the actions also on the basis of the previous observation/internal/action state.

The environmental conditions that are subjected to variation, in the case of this problem, are the initial position of the cart and the initial angular positions and velocity of the poles. These properties vary in the following intervals: $[-1.944 < x < 1.944, -1.215 < \dot{x} < 1.215, -0.0472 < \theta_1 < 0.0472, -0.135088 < \dot{\theta}_1 < 0.135088, -0.10472 < \theta_2 < 0.10472, -0.135088 < \dot{\theta}_2 < 0.135088]$. Evaluation episodes terminate after 1000 steps or prematurely when the angular position of one of the two poles exceeded the range $[-\frac{\pi}{5}, \frac{\pi}{5}]$ rad or the position of the cart exceed the range [-2.4, 2.4] m. During the evolutionary process, the environmental conditions are determined by using 5 values distributed in uniform manner within the intervals described above. Consequently, the number to different environmental conditions that can be experienced by the evolving agents are $5^6 = 15625$.

The cart is rewarded with 1 point every step until the termination of the episode, i.e. it is rewarded for the ability to keep the poles balanced as long as possible. The state of the sensors, the activation of the neural network, the force applied to the cart, and the position and velocity of the cart and of the poles are updated every 0.02 s.

The long double pole differs from the classic non-markovian double-pole balancing problem (Wieland, 1991) for the length of the second pole and the range of variation of initial state of the agent. More specifically, the length and the mass of the second pole corresponds to $\frac{1}{2}$ of the length of the first pole (instead of $\frac{1}{10}$), and the range of variation of the environmental conditions is much larger than in the classic double-pole balancing problem. These variations increase significantly the complexity of the problem (for more details see Pagliuca, Milano and Nolfi, 2018 and Pagliuca and Nolfi, 2019).

The second problem is the **Bipedal walker hardcore** available in the Open AI Gym library (Brockman et al., 2016). The problem consists in controlling a bipedal agent formed by a skull and two two-segments legs for the ability to walk forward in an environment containing obstacles, holes, and stairs. The agent is rewarded for the distance travelled forward, up to a maximum distance of 300, and punished with -100 for falling down.

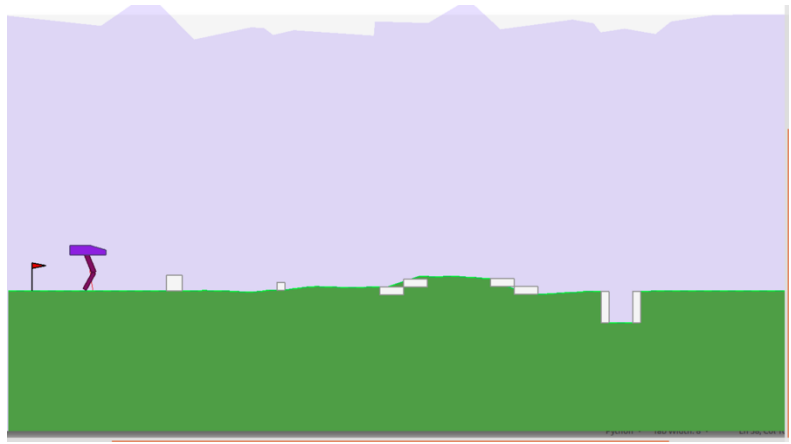

Figure 2. The bipedal worker hardcore problem. The agents is shown on the left of the figure. In this example the environment includes, from left to right, a large stump, a small stump, an uphill stair, a downhill stair, and a hole. The type of each of the five obstacles vary in evaluation episodes.

The agent has 24 sensory neurons that encode the orientation of the skull, the angular, horizontal and vertical velocity of the skull, the position and the velocity of the two hinge joints of each leg, the state of the contact sensors located in the terminal segments the legs, and the state of the ten LIDAR rangefinders that provide information on the nearby frontal portion of the environment. Moreover, the agent has 4 motor neurons that control the torque applied on the joints of the legs.

The neural network of the agent is constituted by a 3-layers feed-forward neural network with 24 sensory neurons, 64 internal neurons, and 4 motor neurons.

The pitfalls present in the environment are subjected to variations. Each environment include a sequence of five obstacles. Since each obstacle is selected among five different alternatives (small stumps, large stumps, uphill stairs, downhill stairs, and holes), the number of possible different environments is $5^5 = 3625$.

Notice that in the case of the long double-pole balancing problem the environmental variations affect the initial position of the agent. In the case of the bipedal walker hardcore, instead, the environmental variations concern the characteristics of the external environment.

The connection weights of the neural network controllers are encoded in a vector of floating point values and evolved through the evolutionary algorithm described in the following section.

**2.2 The evolutionary algorithm extended with curriculum learning**

To evolve the connection weights of the neural network controller we used the Open-AI evolutionary strategy (OpenAI-ES) proposed by Salimans et al. (2017) extended with our curriculum learning algorithm.

We selected the OpenAI-ES method since it is one of the most effective evolutionary methods for continuous control problems (Salimans et al., 2017; Pagliuca, Milano and Nolfi, 2020). The algorithm operates on a population centered on a single parent ($\theta$), uses a form of finite difference method to estimate the gradient of the expected fitness, and update the center of the population distribution with the Adam stochastic optimizer (Kingma & Ba, 2014).

The pseudo-code of the algorithm is reported below. At each generation, the algorithm generates the gaussian vectors $\varepsilon$ that are used to make the offspring, i.e. the perturbed versions of the parent (line 4), which are then evaluated (lines 5-6). The usage of couples of mirrored samples, that receive opposite perturbations, improves the accuracy of the estimation of the gradient (Brockhoff et al., 2010). The offspring are evaluated for $\eta$ episodes in variable environmental conditions (lines 5-6). The average fitness values obtained during evaluation episodes are then ranked and normalized in the range [-0.5, 0.5] (line 7). This normalization makes the algorithm invariant to the distribution of fitness values and reduce the effect of outliers. The estimated gradient g corresponds to the average of the dot product of the samples $\varepsilon$ and of the normalized fitness values (line 8). Finally, the gradient is used to update the parameters of the parent through the Adam (Kingma & Ba, 2014) stochastic optimizer (line 9).

**The OpenAI-ES algorithm**

$\sigma$ = 0.02: mutation variance

$\lambda$ = 20: half population size (total population size = 40)

$\theta$: connection weights

*f()*: evaluation function

optimizer = Adam

$\eta$: number of evaluation episodes

$\varphi$: method used to select the environmental conditions

**1** initialize $\theta_0$
**2 for** *g* = 1, 2, … **do**

**3**   **for** *i* = 1, 2, … $\lambda$ **do**
**4**     sample noise vector: $\varepsilon_i \sim N(0, I)$
**5**     evaluate score: $s_i^+ \leftarrow f(\theta_{t-1} + \sigma * \varepsilon_i, \eta, \varphi)$
**6**     evaluate score: $s_i^- \leftarrow f(\theta_{t-1} - \sigma * \varepsilon_i, \eta, \varphi)$
**7**   compute normalized ranks: u = ranks(s), $u_i \in [-0.5, 0.5]$
**8**   estimate gradient: $g_t \leftarrow \frac{1}{\lambda} \sum_{i=1}^{\lambda}(u_i * \varepsilon_i)$
**9**   $\theta_g = \theta_{g-1}$ + optimizer(g)

In the standard version of the algorithm each offspring is evaluated for $\eta$ episodes in which the environmental conditions are chosen randomly within a uniform distribution. In other words, the function $\varphi$ used to select the environmental conditions correspond to a simple random function. This means that in the case of the long double-pole problem, the position of the cart and the angle and the velocity of the poles are selected randomly with a uniform distribution at the beginning of each evaluation episodes within the variations ranges described in Section 2.1. In the case of the bipedal walker hardcore problem, the type of each of the five obstacles present in the environment is chosen randomly from among one of the five possible types.

In the curriculum learning version of the algorithm, instead, the η environmental conditions experienced during the η evaluation episodes are chosen randomly from among all possible environmental conditions during the first $\frac{1}{10}$ of the evolutionary process only. Then, they are chosen randomly from among η subsets characterized by different levels of difficulty. The subsets are obtained by: (i) normalizing the performance ρ achieved on each environmental condition during the last 5 evaluations in the range [0.0, 1.0], (ii) calculating η performance intervals on the basis of a difficulty function (see below), and (iii) selecting η environmental conditions randomly from η subsets, i.e. selecting η environmental conditions characterized by η different levels of difficulty. See the pseudocode included below.

In case a subset results empty, the environmental conditions are chosen from one of the two nearest non-empty subsets, chosen randomly. The initial phase in which the environmental conditions are chosen randomly from among all possible conditions is necessary in order to start estimating the performance obtained in different conditions.

**Curriculum learning (selection of environmental conditions)**
η: number of evaluation episodes
$K$: environmental condition
κ: selected environmental conditions
ρ: average performance during last 5 evaluation episodes
*f()*: difficulty function

1  **if i <** (tot_iterations / 10)
2   **for** *e* = 1, 2, …η **do**
3    κ[*e*] = rand(*K*)                              # select randomly from all conditions
4  **else**
5   $\hat{\rho} = \frac{\rho - min(\rho)}{max(\rho) - min(\rho)}$        # normalize performance
6   **for** *e* = 1, 2, …η **do**
7    κ[*e*] = rand($X \in K | f(\frac{e-1}{\eta}) \leq \hat{\rho} \leq f(\frac{e}{\eta})$))   # select randomly from η conditions subsets

This technique ensures that each agent experiences conditions with different level of difficulties, ranging from the easiest to the hardest conditions. Moreover, it ensures that all agents experience conditions with a similar overall level of difficulty. It thus reduces the level of stochasticity of the fitness measure caused by the fact that the fitness of lucky and un-lucky individuals, i.e. of individuals that encountered easier and harder environmental conditions, tend to be over-estimated and under-estimated, respectively. The fact that the selection of the environmental conditions is stochastic reduces the risk of overfitting and ensures that the estimated difficulty of the environmental conditions continue to be updated during the course of the evolutionary process. Indeed, all environmental conditions have a chance to be selected, although the probability with which they are selected varies. For an analysis of the importance of using stochasticity to avoid overfitting in a related model see Yu, Turk & Liu (2018).

The need to avoid overfitting is also the reason why we preferred our method, that operates by selecting the environmental conditions within the set of all possible conditions, to possible alternative methods that operate by generating suitable environmental conditions. An approach of this type can be effective to generate conditions characterized by few variables, e.g. to generate the goal of an agent trained through reinforcement learning (Held et al., 2017). However, it might not be suitable to generate sufficiently varied conditions characterized by several dimension of variability.

The overall level of difficulty can be varied by varying the function used to determine the intervals of the categories. For this reason, we will refer to this function with the term difficulty function.

The utilization of linear function (Figure 4, blue line) does not alter the overall level of the difficulty of the η environmental conditions experienced by the agents with respect to a standard evolutionary algorithm in which the environmental conditions are selected randomly within all possible conditions.

Instead, the utilization of a power function permits to select preferentially difficult environmental conditions (Figure 4, red line). Indeed, the intervals calculated with a power function are larger for easy conditions, i.e. conditions in which the agents obtained a high fitness, and progressively smaller for difficult conditions. i.e. conditions in which the agents achieved a low fitness (Figure 4, red line). This implies that the numerosity of the categories is large for the easiest conditions and progressively smaller for the more difficult conditions and consequently that the conditions belonging to the easier subsets are chosen less frequently than the conditions belonging to the harder subsets. Notice that, as mentioned above, the performance of the environmental conditions is calculated on the basis of the fitness obtained by recent evolving agents. Consequently, the performance values do not indicate an absolute level of difficulty but rather the level of difficulty relative to the skills of the current evolving agents.

The intensity of the preferential selection of difficult condition can be varied by varying the exponential of the power function. The higher the exponential of the function is, the higher the overall difficulty of the selected environmental conditions becomes.

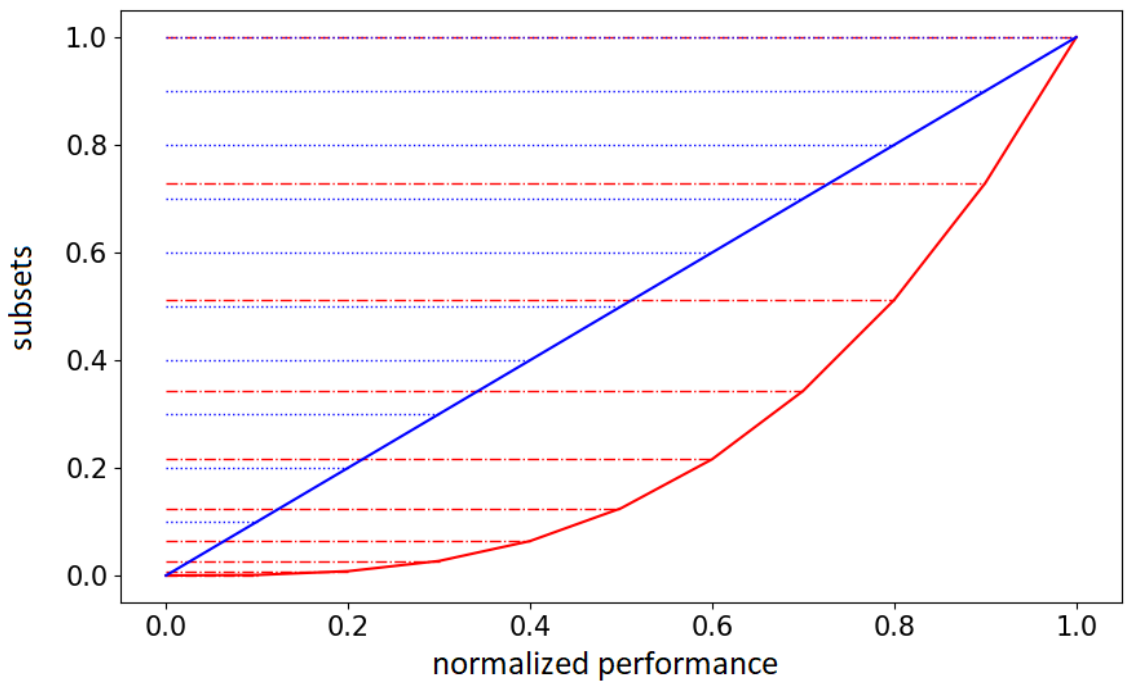

Figure 4. Difficulty intervals obtained by using a linear and a quadratic power difficulty function. Data shown in blue and red, respectively. The horizontal axes represent performance normalized in the range [0.0, 1.0], i.e. the fitness obtained by the last 5 agents normalized in the range [0.0, 1.0]. The vertical axes represents the intervals of the subsets, with η = 10. The intervals of the subsets in the case of the linear function are [0.0, 0.1, 0.2, 0.3, 0.4, 0.5, 0.6, 0.7, 0.8, 0.9, 1.0]. The intervals in the case of the quadratic power function are [0.0, 0.01, 0.04, 0.09, 0.16, 0.25, 0.36, 0.49, 0.64, 0.81, 1.0].

## 3. Results

Figure 5 shows the results obtained on the long double-pole problem by using the standard algorithm and the algorithm with curriculum learning. For the latter case, we report the results obtained with the linear difficulty function and with power difficulty functions with exponentials 2, 3 and 4. The evolutionary process has been continued for $1 \cdot 10^9$ evaluation steps. Performance refer to the average fitness obtained by the best agents of each replication post-evaluated for 1000 evaluation episodes during which the environmental conditions were chosen randomly with a uniform distribution within the range described in Section 2.1.

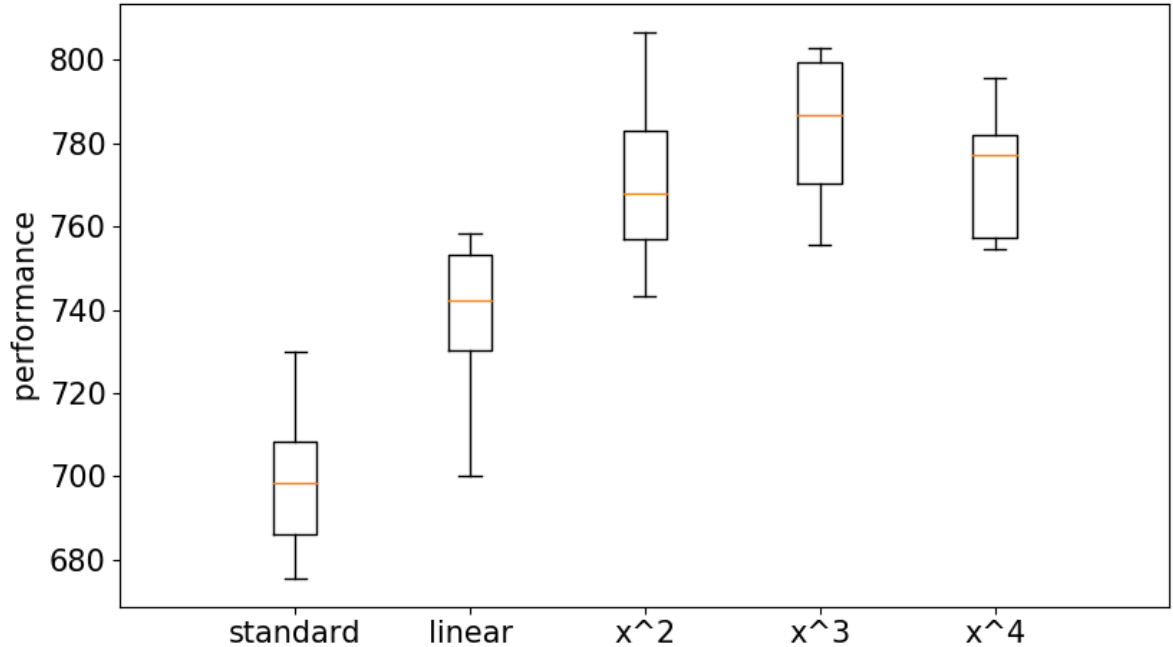

Figure 5. Performance of the agents evolved on the the double pole balancing problem with the standard algorithm and with the curriculum learning algorithm. For the latter algorithm the figure reports the results obtained by using a linear difficulty function (linear) and a power function with exponential 2, 3 and 4 (x^2 x^3, and x^4). Each boxplot shows the results obtained in 10 replications. Boxes represent the inter-quartile range of the data and horizontal lines inside the boxes mark the median values. The whiskers extend to the most extreme data points within 1.5 times the inter-quartile range from the box.

The group comparison performed with the Kruskal-Wallis test indicates that at least one condition dominates other conditions (data number = 10, p-value < 0.001). The pairwise comparison carried out with the Mann-Whitney U test with Bonferroni correction shows that the curriculum learning with the power functions outperform the curriculum learning condition with the linear function (data number = 10, p-value < 0.001 with Bonferroni corrections α = 5) and the standard method (data number = 10, p-value < 0.001 with Bonferroni corrections α = 5). The curriculum learning with the linear function outperform the standard condition (data number = 10, p-value < 0.01 with Bonferroni corrections α = 5). The curriculum learning condition with the cubic power function outperform the curriculum learning conditions with the x^2 and x^4 power functions (data number = 10, p-value < 0.01 with Bonferroni corrections α = 5). See also Table 1.

The fact that the curriculum learning algorithm with the linear difficulty function outperforms the standard algorithm demonstrates that a first advantage of the curriculum learning algorithm is due to its ability to expose all agents to environmental conditions with similar levels of difficulty.

The fact that the curriculum learning algorithm with power difficulty functions outperform the curriculum learning algorithm with the linear difficulty function demonstrates that selecting preferentially difficult environmental conditions promotes the evolution of better solutions.

The analysis of the performance during the evolutionary process indicates that the curriculum learning algorithm with the cubic difficulty function starts to outperform the standard algorithm rather early (Figure 6).

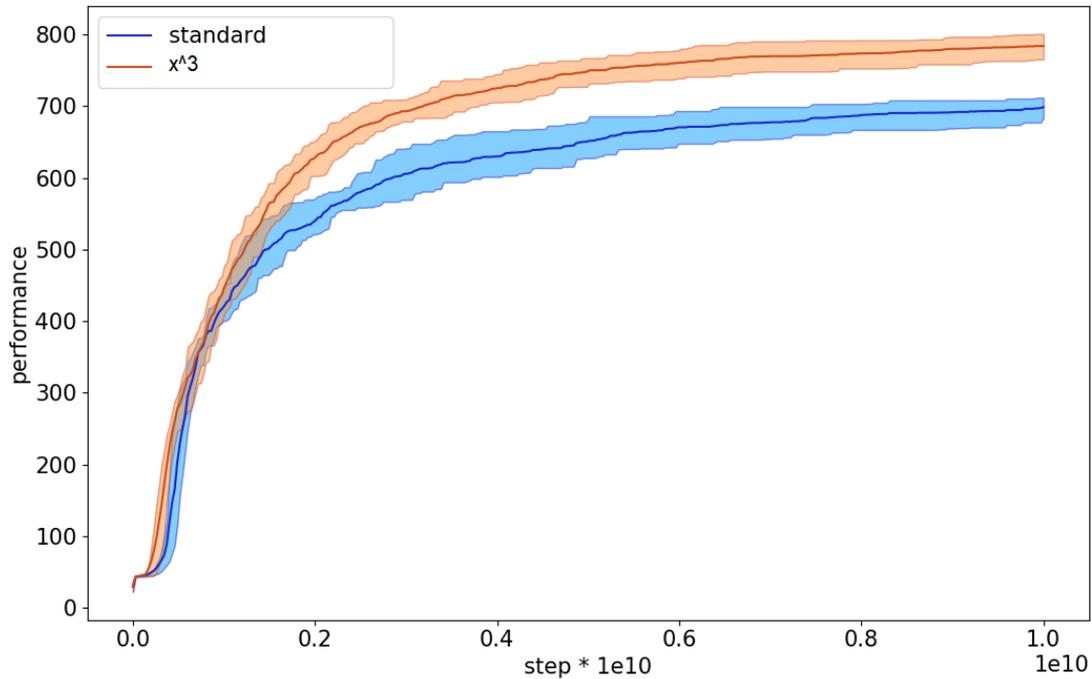

Figure 6. Performance of the experiments performed with the standard algorithm and with the curriculum learning algorithm with the cubic difficulty function during the evolutionary process. Each curve shows the average results of 10 replications. The shadow indicate the 90% bootstrapped confidence intervals.

Figure 7 shows an analysis of the performance across generations for different environmental conditions. For this analysis we considered all the possible environmental conditions that can be generated by combining 3 values, distributed in a uniform manner within the intervals described above, for the 6 environmental variables subjected to variation, i.e. the initial position and velocity of the cart and the initial angular positions and velocity of the poles. Consequently, the number to different environmental conditions on which the agents are post-evaluated are $3^6 = 729$. The comparison of performance obtained with the standard algorithm (Figure 7, top) and with the curriculum learning algorithm with the cubic difficulty function shows that the latter algorithm equals or outperforms the former algorithm in the large majority of the environmental conditions (Figure 7, bottom).

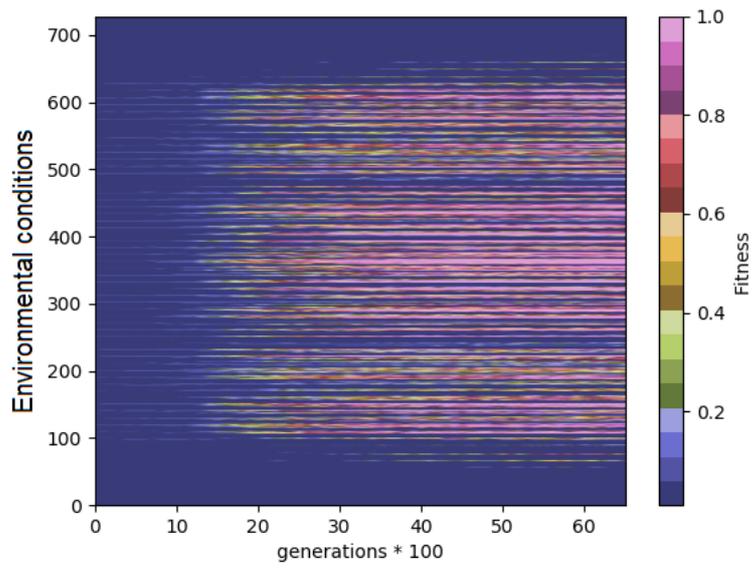
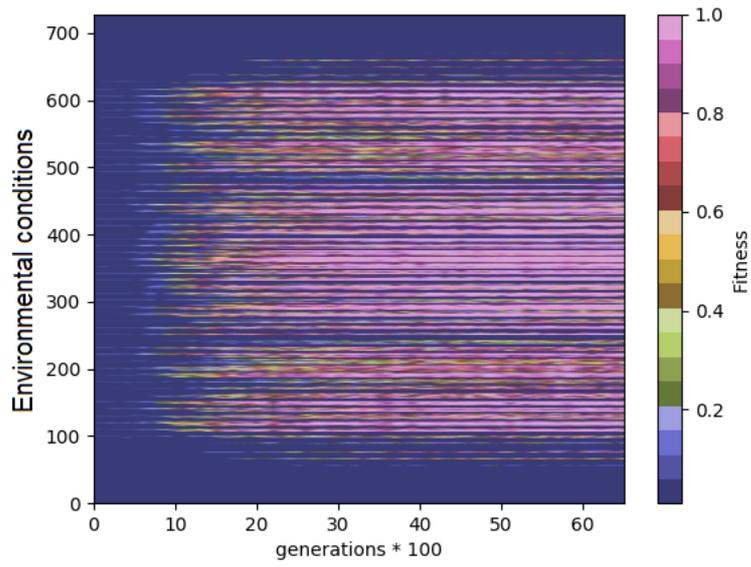
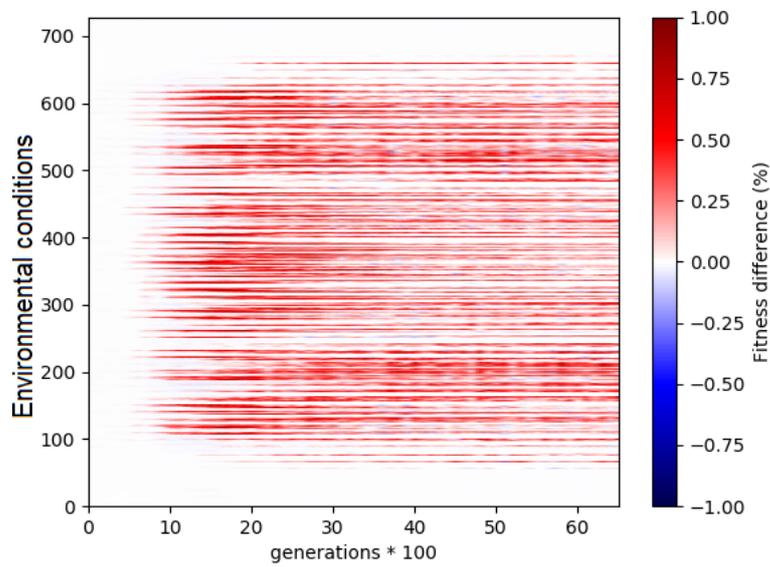

Figure 7. Heatmap of the performance for different environmental conditions during the evolutionary process. The vertical and horizontal axes indicate the environmental conditions and the generations, respectively. The top and central figures show the performance obtained in the experiment performed with the standard algorithm and with the curriculum learning algorithm with the cubic difficulty function. The bottom figure shows the difference between the performance obtained in the two conditions. Average results of 10 replications for each experimental condition.

Figure 8 shows the performance obtained in the case of the bipedal walker hardcore problem. The evolutionary process has been continued for $3 \cdot 10^8$ evaluation steps. Performance refer to the average fitness obtained by the best agents of each replication post-evaluated for 500 evaluation episodes during which the category of each of the five obstacles has been selected randomly. Also in the case of this problem the Kruskal-Wallis test indicates that at least one condition dominates other conditions (data number = 10, p-value > 0.05). The Mann-Whitney U pairwise comparison shows that the curriculum learning with the power functions outperform the curriculum learning condition with the linear function (data number = 10, p-value < 0.01 with Bonferroni corrections α = 5) and the standard method (data number = 10, p-value < 0.01 with Bonferroni corrections α = 5). The curriculum learning conditions with the $x^2$, $x^3$ and $x^4$ power functions do not differ among themselves (data number = 10, p-value > 0.05 with Bonferroni corrections α = 5). See also Table 2.

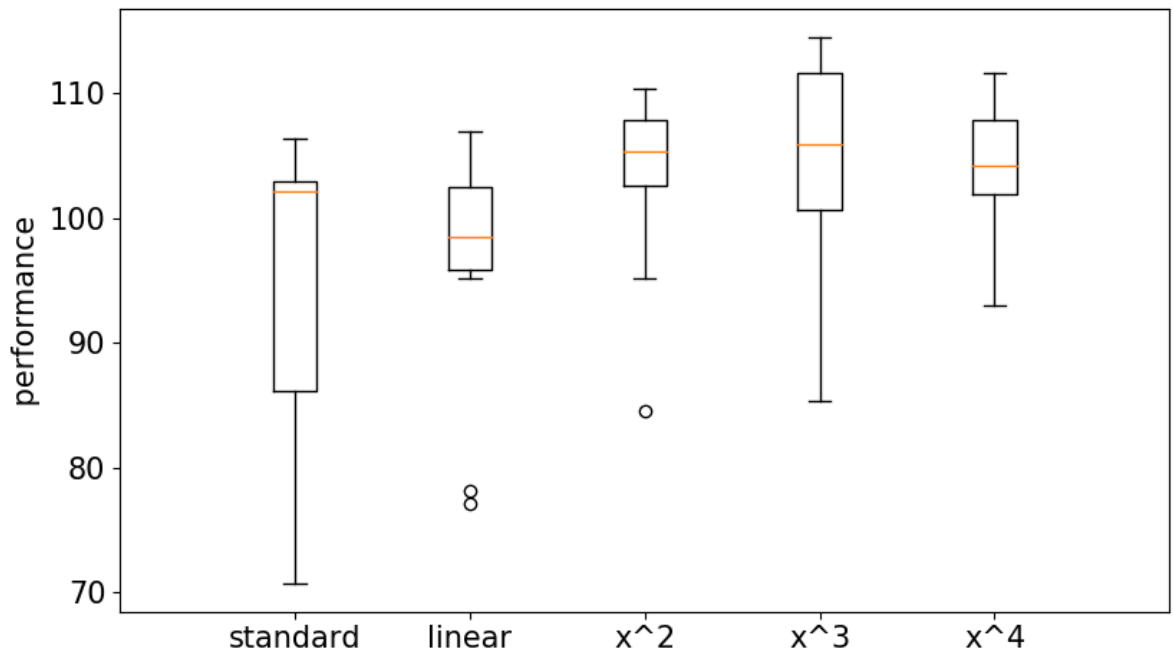

Figure 8. Performance of the agents evolved on the bipedal hardcore problem with the standard algorithm and with the curriculum learning algorithm with the linear difficulty function (linear) and a power function with exponential 2, 3 and 4 ($x^2$ $x^3$, and $x^4$). Each boxplot shows the results obtained in 10 replications. Boxes represent the inter-quartile range of the data and horizontal lines inside the boxes mark the median values. The whiskers extend to the most extreme data points within 1.5 times the inter-quartile range from the box.

The analysis of the course of the evolutionary process indicates that the curriculum learning algorithm with the cubic difficulty function starts to outperform the standard algorithm rather early (Figure 9).

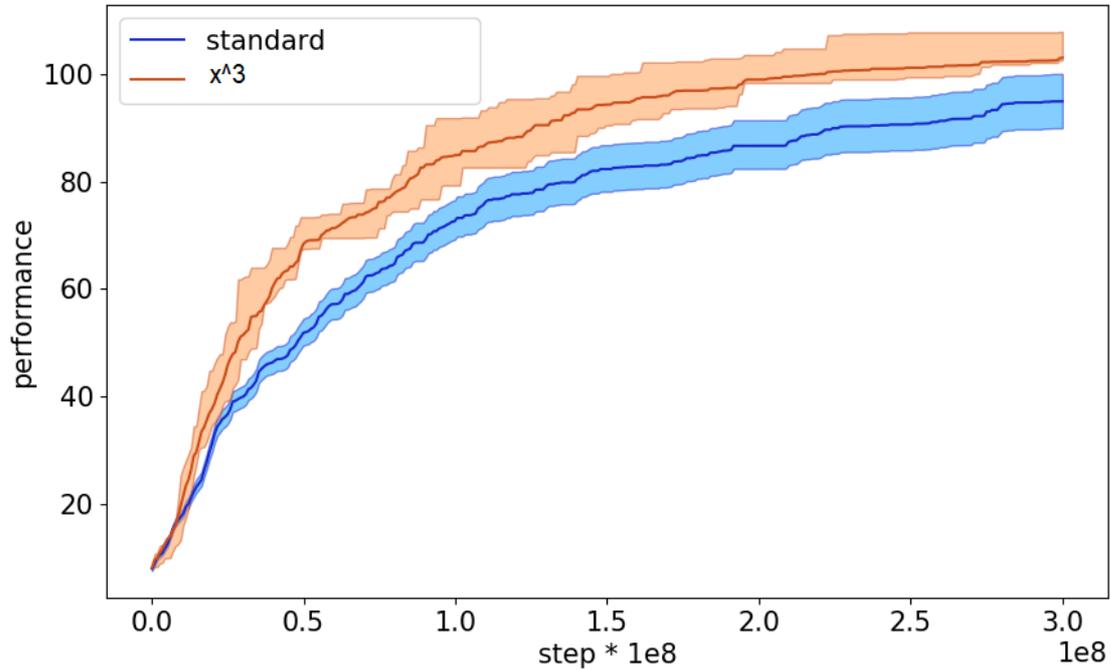

Figure 9. Performance of the experiments performed with the standard algorithm and with the curriculum learning algorithm with the cubic function during the evolutionary process. The curves show the average results of 10 replications. The shadow indicate the 90% bootstrapped confidence intervals.

Figure 10 shows an analysis of the performance across generations for different environmental conditions. For this analysis we considered all the possible environmental conditions that can be generated by combining all the 5 obstacles. Consequently, the number of different environmental conditions on which the agents are post-evaluated are $5^5 = 3625$. Also in this case, the comparison of performance obtained with the standard algorithm (Figure 10, top) and with the curriculum learning algorithm with the cubic function (Figure 10, center) shows that the latter algorithm equals or outperform the former algorithm in the large majority of the environmental conditions (Figure 10, bottom).

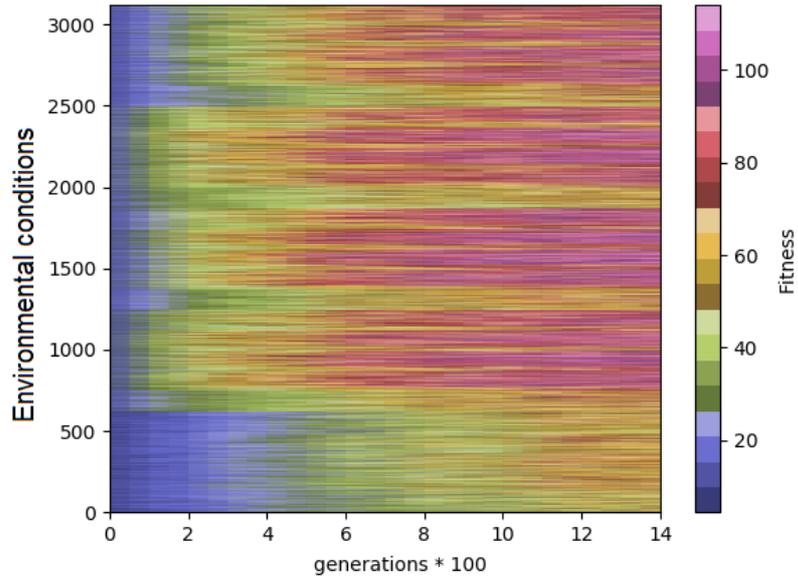

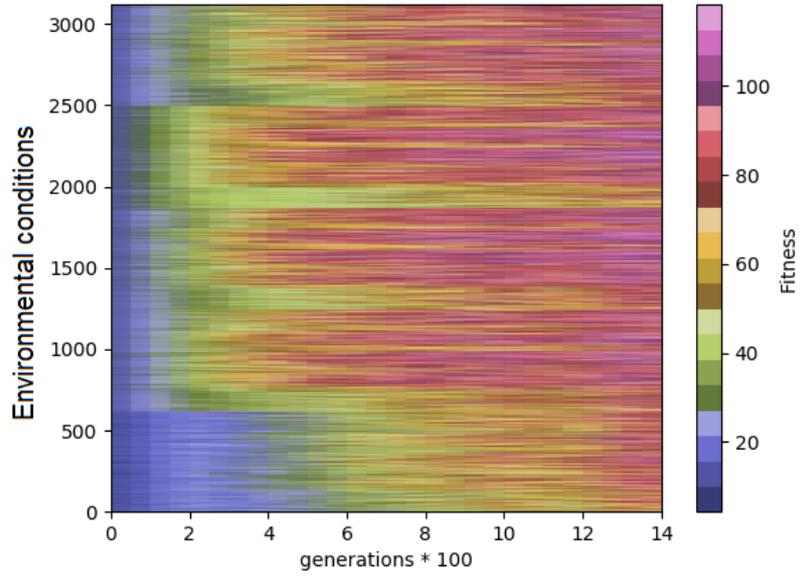

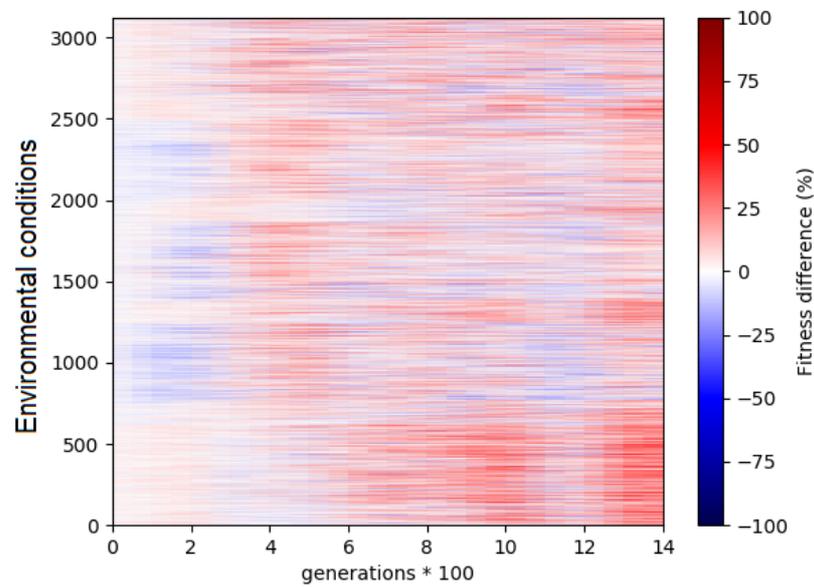

Figure 10. Heatmap of the performance for different environmental conditions during the evolutionary process. The vertical and horizontal axes indicate the environmental conditions and the generation. The top and central figures show the performance obtained in the experiment performed with the standard algorithm and with the curriculum learning algorithm with the cubic difficulty function. The bottom figure shows the difference between the performance obtained in the two conditions. Average results of 10 replications for each experimental condition.

## 4. Discussion

We proposed a method that enables evolutionary algorithms to select the environmental conditions that facilitate the evolution of effective solutions. This is realized by adding a curriculum learning component that estimates the difficulty level of the environmental conditions from the perspective of the evolving agents and selects conditions with different level of difficulties in which the frequency of difficult cases is greater than the frequency of easier cases.

The estimation of the difficulty level of the environmental conditions is performed on the basis of the fitness obtained in those conditions by agents evaluated recently. The selection of suitable environmental conditions is realized by selecting η environmental conditions from η corresponding subsets characterized by different levels of difficulty, where η is the number of evaluation episodes. Finally, the preferential selection of difficult conditions is realized by increasing the probability to select difficult environmental conditions, i.e. by determining the intervals of the subsets with a power function. The utilization of this method also reduces the stochasticity of the fitness measure since it ensures that agents are exposed to environmental conditions that have similar levels of difficulty.

The curriculum learning component proposed is general and can be combined with any evolutionary algorithm. In this paper, we verified its efficacy in combination with the Open-AI neuro-evolutionary strategy (Salimans et al., 2017), one of the best algorithms available at the state of the art. We evaluated the efficacy of the method on the long double-pole and on the bipedal hardcore walker problems, that are commonly used to benchmark evolutionary and reinforcement learning problems and that require to handle significantly varying environmental conditions.

The obtained results indicate that the curriculum learning method produces significantly better results in the two problems considered in all conditions. The fact that the curriculum learning condition with the cubic difficulty function produce significantly better performance than the curriculum learning condition with the linear function confirms that the preferential selection of difficult condition enhances the efficacy of the evolutionary process. The fact that the curriculum learning condition with the linear difficulty function outperforms the standard algorithm demonstrates that exposing the agents to environmental conditions with similar level of difficulty also enhances the efficacy of the evolutionary process. Consequently, the advantage gained by the curriculum learning with the cubic difficulty function can be ascribed both to the ability of the method to reduce the stochasticity of the fitness measure and to the ability to increase the frequency of difficult environmental conditions.

The method proposed presents several advantages with respect to related techniques such as incremental evolution and competitive coevolution: it can be applied to any problem, it does not require to set additional hyperparameters, and it does not require the usage of domain specific knowledge. A limit of the method concerns its scalability with respect to the number of environmental variables that are subjected to variations. This since the time required to estimate the difficulty level of the environmental conditions increases exponentially with the number of variables subjected to variation. This is a general challenge that also affect incremental evolution and competitive evolution. A possible solution to this problem, that deserves to be investigated in future studies, consists in estimating the difficulty level of the environmental conditions through a neural network that receives as input the environmental conditions and produce as output the estimated difficulty levels. This network could be trained on the basis of the same historical data that we used in our method. The potential advantage of this approach is that it can generalize, i.e. it can estimate correctly the difficulty level also of environmental conditions that were not experienced in previous evaluations.

Another aspect that can be considered in future studies is the criterion used to select the environmental conditions. In this work we proposed an approach that relies on in the difficulty level of the environmental conditions. An alternative criterion, that has been explored in the context of supervised learning (see Graves et al., 2017; Matiisen et al., 2019) and reinforcement learning (see Portelas et al., 2020) is learning progress, namely the propensity of examples or of environmental conditions to induce learning progress.

## 5. Appendix

Table 1 and 2 report the p-value for the pairwise comparison among all the conditions in the case of the long double-pole and bipedal walker hardcore problems, respectively.

Table 1. P-values from the Mann-Whitney U test with Bonferroni correction α = 5 for the long double-pole problem.

|          | standard | linear | $X^2$  | $X^3$      | $X^4$      |
|----------|----------|--------|--------|------------|------------|
| Standard | -        | 0.008  | 0.004  | $2*10^4$   | $7*10^4$   |
| Linear   | -        | -      | 0.007  | $8*10^4$   | 0.004      |
| $X^2$    | -        | -      | -      | 0.005      | 0.08       |
| $X^3$    | -        | -      | -      | -          | 0.006      |
| $X^4$    | -        | -      | -      | -          | -          |

Table 2. P-values from the Mann-Whitney U test with Bonferroni correction α = 5 for the bipedal walker hardcore problem.

|          | standard | linear | $X^2$  | $X^3$  | $X^4$  |
|----------|----------|--------|--------|--------|--------|
| Standard | -        | 0.007  | 0.005  | 0.004  | 0.003  |
| Linear   | -        | -      | 0.007  | 0.006  | 0.006  |
| $X^2$    | -        | -      | -      | 0.06   | 0.08   |
| $X^3$    | -        | -      | -      | -      | 0.1    |
| $X^4$    | -        | -      | -      | -      | -      |